\documentclass{article}

\usepackage[dblblindworkshop, final]{neurips_2025}
\workshoptitle{Interpreting Cognition in Deep Learning Models}

\usepackage[utf8]{inputenc} 
\usepackage[T1]{fontenc}    
\usepackage{hyperref}       
\usepackage{url}            
\usepackage{booktabs}       
\usepackage{amsfonts}       
\usepackage{nicefrac}       
\usepackage{microtype}      
\usepackage{xcolor}         

\usepackage{soul}
\usepackage{hyperref} 
\usepackage{graphicx}
\usepackage{amsmath}
\usepackage{cleveref}
\usepackage{xspace}

\newcommand{\allSubjectsBrainWidth}{0.45\textwidth}

\newcommand{\windowLength}{T}

\newcommand{\Friends}{\emph{Friends}\xspace}
\newcommand{\Mustard}{MUSTtARD\xspace}

\newcommand{\TVLT}{TVLT\xspace}

\newcommand{\significantAsterisk}{\textcolor{red}{*\xspace}}
\newcommand{\highlySignificantAsterisk}{\textcolor{red}{**\,\xspace}} 

\title{The One Where They Brain-Tune for Social Cognition: Multi-Modal Brain-Tuning on \emph{Friends}}

\author{%
    Nico Policzer$^{1,2,3}$ \quad Cameron Braunstein$^{1, 2}$ \quad Mariya Toneva$^2$ \\
    $^1$Saarland University, Saarbrucken, Germany \\
    \quad $^2$ MPI for Software Systems, Saarbrucken, Germany \\
    \quad $^3$ University of British Columbia, Vancouver, Canada\\
    \texttt{npoliczr@student.ubc.ca}\\
    \texttt{braunstein@cs.uni-saarland.de}\\
    \texttt{mtoneva@mpi-sws.org}
}
\begin{document}

\maketitle

\begin{abstract}
Recent studies on audio models \cite{moussa2025improvingsemanticunderstandingspeech, tuningfreteault} show \emph{brain-tuning}--fine-tuning models to better predict corresponding fMRI activity--improves brain alignment and increases performance on downstream semantic and audio tasks. We extend this approach to a multimodal audio-video model to enhance social cognition, targeting the Superior Temporal Sulcus (STS), a key region for social processing, while subjects watch \emph{Friends}. We find significant increases in brain alignment to the STS and an adjacent ROI, as well as improvements to a social cognition task related to the training data--sarcasm detection in sitcoms. In summary, our study extends brain-tuning to the multi-modal domain, demonstrating improvements to a downstream task after tuning to a relevant functional region.
\end{abstract}

\section{Introduction}
\label{sec:intro}

Recent works in fine-tuning audio models to human fMRI data, specifically language and auditory areas, show improvements to brain alignment, as well as increases to performance on semantic and audio evaluations \cite{tuningfreteault, moussa2025improvingsemanticunderstandingspeech, negi2025brain}. However, frontier AI models are increasingly multi-modal \cite{caffagni2024revolutionmultimodallargelanguage,multimodal_learning_survey}. These models are uniquely posed to model human social cognition, \emph{i.e.}, inferring a perceived person's internal state, which requires integrating information across modalities \cite{campanella2007integrating, benetti2023multimodal, bae2016pt735} and is critical as AI becomes more integrated in our daily lives \cite{bolotta2022social}. However, a recent study \cite{garcia2025modeling} identified a major gap in AI models' abilities to match human social perception, as well as encode brain activity in the lateral stream, a processing stream proposed for social cognition \cite{pitcher2021lateralevidence}. The Superior Temporal Sulcus (STS), the end point of the lateral stream, is a brain-region that has been shown to encode features of social interaction relevant to social cognition \cite{masson2021functional, isik2017social, walbrin2018neural, pitcher2021lateralevidence, hein2008superior, allison2000social, deen2015functional}. We therefore investigate whether brain-tuning an audio-video model to the STS can 1) improve brain encoding of the STS and other lateral stream ROIs, and 2) increase downstream performance on social cognition tasks.

Concretely, we brain-tune the joint audio-video transformer model, TVLT \cite{tang2022tvlt}, to the STS using data from n=6 subjects from the Courtois Neuromod Dataset \cite{boyle2020courtois}, while subjects watch the sitcom \emph{Friends}. This significantly increases alignment to both the STS (our tuning target), and a further (non-targetted) lateral-stream ROI.

To evaluate social cognition, we first test whether tuning improves performance in a context similar to the \Friends training data, and report significantly increased performance on a sarcasm detection task containing data from sitcoms (MUStARD). We then test whether these improvements generalize to a social cognition task in a markedly different context, emotion and sentiment prediction on CMU-MOSEI, but find no significant increase in performance from brain-tuning, suggesting that tuning improves social cognition performance in related contexts but does not generalize to contexts not represented in training.

Our main contributions are as follows: We extend the brain-tuning methodology to a multi-modal audio-video domain, and show, for the first time, that brain-tuning a model to an ROI involved in social cognition can improve its performance on a related social cognition task.
This provides further evidence
\cite{moussa2025improvingsemanticunderstandingspeech, tuningfreteault} that targeted brain tuning to specific functional ROIs can increase alignment and improve performance to related downstream tasks.

\section{Related Work}
\textbf{Lateral Stream \& Superior Temporal Sulcus.} The lateral stream has been recently proposed as a third visual processing stream specialized for dynamic social processing (\cite{pitcher2021lateralevidence}), in addition to the classical ventral and dorsal streams. Its endpoint, the Superior Temporal Sulcus (STS), 
robustly encodes features of social interaction, allowing for the processing of the intentions and inner states of others. \cite{masson2021functional, isik2017social, walbrin2018neural, pitcher2021lateralevidence, hein2008superior, allison2000social, deen2015functional}. This motivates our use of STS activity as a tuning signal for social cognitive tasks. See the appendix for a visualization of the STS on the whole brain.

\textbf{Prior work in Brain Alignment and Brain Tuning.} There is a large body of work measuring brain alignment in neural models \cite{oota2025correlating, oota2025multimodalbrainencodingmodels, dong2023visionlanguageintegrationmultimodalvideo, nakagi-etal-2024-unveiling,lu2022multimodalfoundationmodelsbetter}, however, few \cite{moussa2025improvingsemanticunderstandingspeech, tuningfreteault} studies fine-tune a pretrained model to increase alignment. Unlike these prior works \cite{moussa2025improvingsemanticunderstandingspeech, tuningfreteault} which fine-tune audio-only models to late language regions and evaluate on downstream auditory and semantic tasks, we instead tune our multi-modal model to the STS, and evaluate on downstream social cognition tasks. Our work differs from recent multi-modal brain encoder work \cite{dascoli2025tribetrimodalbrainencoder}, which trains a dedicated deep network for brain prediction across regions. In contrast, we tune an existing model to a specific functional region, and aim to improve alignment to that region and performance on a related task.

\section{Method}\label{method}
\begin{figure}
    \centering
    \includegraphics[width=1\linewidth]{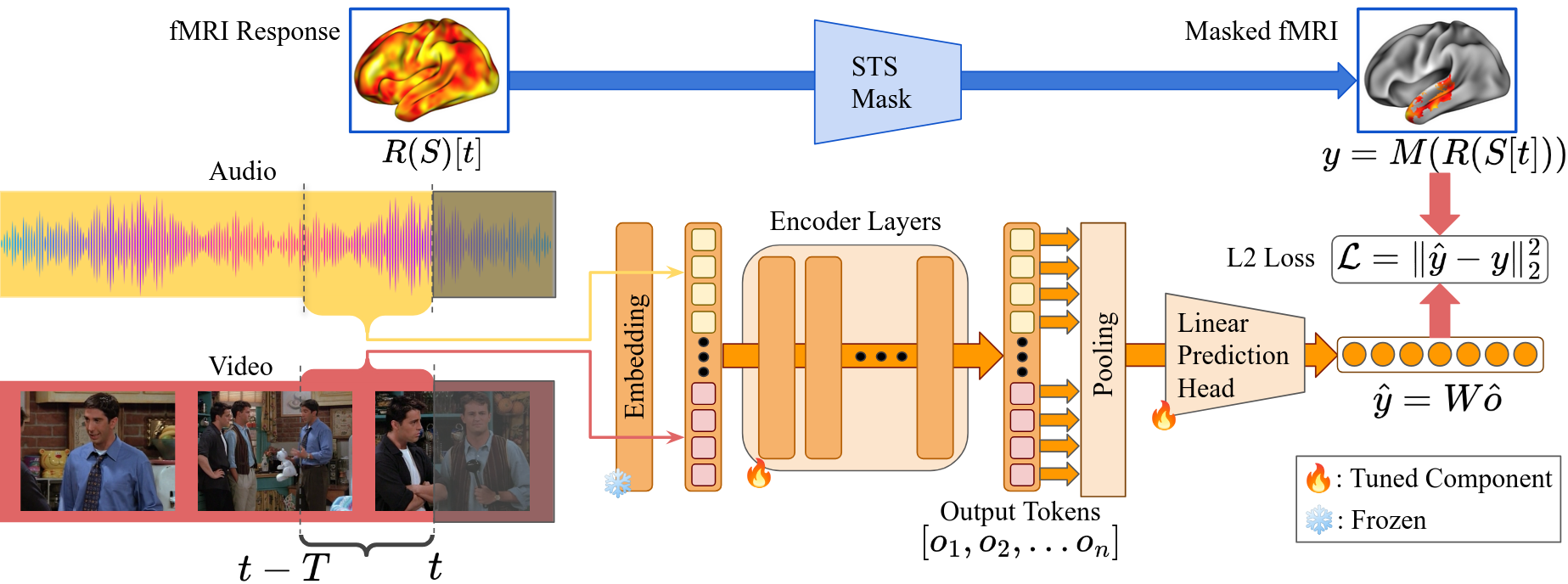}
    \caption{Our audio-video brain-tuning approach. Audio-video stimuli are perceived by the subject, and input to the model, and we fine-tune the model and projection head to better predict corresponding brain activation.}
    \label{fig:brain_tuning_method}
\end{figure}
\subsection{Model and Stimulus}\label{subsec:model_and_stimulus}

\textbf{Model Selection.} Recent works in Video-Language multimodal models are broadly split into LLM-based methods (\cite{hu2024minicpmunveilingpotentialsmall,liu2025olapushingfrontiersomnimodal,li2025baichuanomni15technicalreport,shu2023audiovisualllmvideounderstanding,fu2025vitaopensourceinteractiveomni,sun2024videosalmonnspeechenhancedaudiovisuallarge,cheng2024videollama2advancingspatialtemporal,xu2025qwen25omnitechnicalreport,lyu2023macawllmmultimodallanguagemodeling}) and feature encoder-based methods (\cite{zellers2022merlotreserveneuralscript,girdhar2023imagebind,tang2022tvlt}). We chose to tune the  \textit{Textless Vision Language Transformer} (TVLT) \cite{tang2022tvlt}, due to architectural similarities with the models brain-tuned in \cite{moussa2025improvingsemanticunderstandingspeech} including number of encoders layers (12), embedding size (768), and total number of parameters ($ \sim 90$M). It is pretrained on around 130K hours of audio-video with a joint masked auto-encoding and vision-audio matching objective. An initial embedding layer embeds each 16×16 patch of each video frame, and converts the audio to a log-mel spectrogram, which are then jointly encoded through the transformer layers.

\textbf{fMRI Data.} We use a subset of the preprocessed fMRI data from the 2022-alpha release of the Courtois Neuromod Dataset \cite{boyle2020courtois}, containing n=6 subjects  watching seasons 1-4 of the sitcom \textit{Friends} (seasons 1-3 for training, 4 for evaluation). It is one of the largest available fMRI dataset of participants watching audio-video stimuli, and has previously been used for brain-tuning an audio model \cite{tuningfreteault}. More information about this dataset can be found in the appendix. 

\textbf{Cross Subject Prediction Accuracy Estimation.} Noise in the fMRI data--both natural fMRI noise as well as signal unrelated to the stimulus--can impair both our brain-tuning and evaluation procedures. To estimate the level of noise present in each voxel, we follow recent studies \cite{oota2025multimodalbrainencodingmodels}, \cite{dong2023visionlanguageintegrationmultimodalvideo} in adapting \cite{schrimpf2021neural}'s method to estimate cross-subject prediction accuracy for each voxel. See \ref{appendix:noise_ceilings} for technical details. Following previous brain-tuning studies \cite{moussa2025improvingsemanticunderstandingspeech}, we filter out voxels with a low cross-subject prediction accuracy to tune only on voxels reliably related to the stimulus. We attempt to reach the threshold of 0.4 used in prior brain-tuning \cite{moussa2025improvingsemanticunderstandingspeech}, but find that beyond a threshold of $0.25$, all STS voxels are removed for some subjects, preventing training (see \cref{appendix:noise_ceilings}). Therefore, we set our threshold to $0.25$, leaving subjects with 100-700 STS brain-tuning target voxels.
We also use cross-subject prediction accuracy to normalize voxel activations when computing normalized brain alignment (more in \cref{evaluation-proc}).
 
\subsection{Brain Tuning}\label{subsec:braintuning} 
\textbf{Training Objective.} Following \cite{moussa2025improvingsemanticunderstandingspeech}, we fine-tune our pretrained model to predict the fMRI voxels in the STS with a high cross subject prediction accuracy. Formally, let $S$ be the synchronized audio–video stimulus, and $R(S)[t]$ the recorded fMRI response at time $t$. 
We define a voxel masking function $M$ such that:
\begin{equation}
y = M(R(S)[t]),
\end{equation}
where $y \in \mathbb{R}^m$ is the STS-masked fMRI vector of $m$ voxels.
Let $T$ be the length of the temporal receptive field, approximately 12s in our case. 
We take an audio–video clip from $t - T$ to $t$, denoted $S[t-T : t]$, and process it with TVLT to obtain output tokens $[o_1, o_2, \dots, o_n] \in \mathbb{R}^{n \times 768}$.
We mean-pool the tokens: $\hat{o} = \frac{1}{n} \sum_{i=1}^n o_i$.
A linear projection layer $W \in \mathbb{R}^{m \times 768}$ maps $\hat{o}$ to the predicted fMRI vector:
\begin{equation}
\hat{y} = W \hat{o}.
\end{equation}
We minimize the L2 loss, $\mathcal{L}$, between the predicted voxel activations $\hat{y}$ and true activations $y$:
\begin{equation}
\mathcal{L} = \|\hat{y} - y\|^2_2,
\end{equation}
and backpropagate $\mathcal{L}$ through both the projection layer $W$ and the TVLT transformer layers. 
The overall process is illustrated in \cref{fig:brain_tuning_method}.

\textbf{Training Details.} To predict each fMRI snapshot, we give the model the previous 8 TR-lengths ($\windowLength = 11.92$s) of audio-video stimulus. This finite response window is similar to that used in prior work \cite{moussa2025improvingsemanticunderstandingspeech}, and is in line with  the average hemodynamic response cycle of 12s \cite{voss_2016}. Following \cite{tuningfreteault}, we train our model on the first three seasons (68,063 TRs, TR=1.49s) of \Friends, and evaluate on season four.  
Following the finding by \cite{tuningfreteault} that individual models often outperformed models tuned to multiple subjects at once, we tune one model to each subject's (n=6) brain activity.
Due to compute limits, we restrict our tuning to 10 epochs. For each 11.92s clip, we evenly sample $8$ frames from the video following \cite{tang2022tvlt}, and sample audio at the standard 44,100 Hz. We optimize with Adam with a constant learning rate of $1.0 \times 10^{-6}$. Brain-tuning each model uses 1 H100 GPU and 16 AMD EPYC 9654 CPUs on 244 GB of RAM, and takes approximately $70$ hours on an H100 GPU. Each evaluation uses identical compute specs, and takes approximately 90 minutes.

\begin{figure}
    \centering
    \begin{tabular}{cc}
    a)\label{fig:brain_alignment_bar}
    \includegraphics[width=0.4\linewidth]{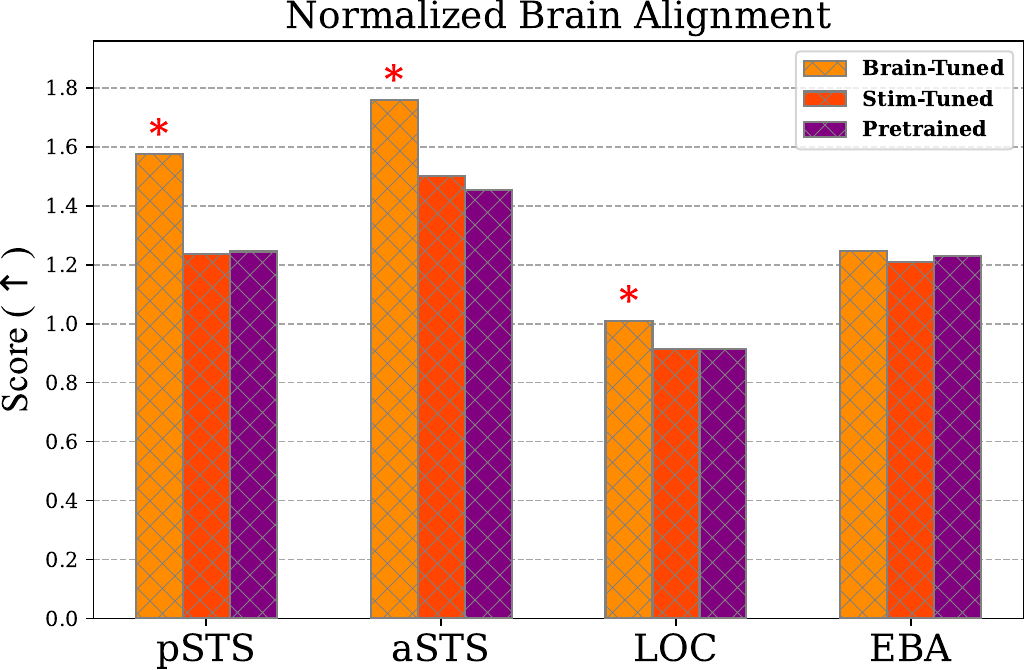} &
    b)
    \raisebox{0.13\height}
    {\includegraphics[width=0.5\linewidth]{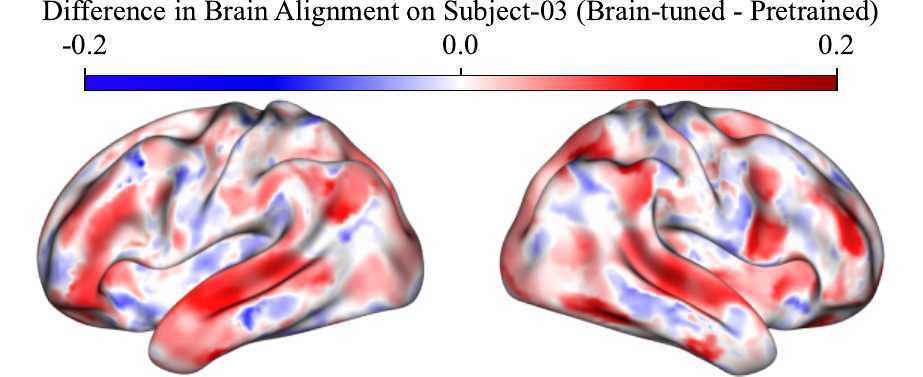}}
    \end{tabular}
    \caption{\textbf{a:} Average change in alignment to lateral ROIs after brain-tuning over subjects. 
    We find significant increases in the pSTS, aSTS, and LOC. 
    \textbf{b:} Change in alignment before and after tuning on Subject-03. Differences for all subjects can be found in the appendix. 
    }
    \label{fig:brain_tuning_single_subject}
\end{figure}
\subsection{Evaluation Procedure}\label{evaluation-proc}

\textbf{Comparison Models.} Following \cite{moussa2025improvingsemanticunderstandingspeech}, we compare against a stimulus-tuned and a pretrained baseline on both brain-encoding and downstream evaluations. The pretrained baseline is the original pretrained TVLT model introduced in \cite{tang2022tvlt}. The stimulus-tuned baseline is trained using the original TVLT joint training objective, with the same video data and learning hyperparameters as the brain-tuned model. This baseline tests whether changes in performance are the result of simply training on the Friends dataset, or are due to the fMRI training objective used in brain-tuning.

\textbf{Encoding Evaluation. } Following \cite{moussa2025improvingsemanticunderstandingspeech}, we use standard voxel-wise encoding models (\cite{antonello2023scaling}
\cite{oota2025multimodalbrainencodingmodels}, \cite{oota2025correlating}) to evaluate the change in brain alignment between our brain-tuned and baseline models.
We follow the same steps as during brain tuning to create TR-video pairs where each fMRI TR is paired with the 8 TR-lengths (11.92s) of video that precede it. This video is input into each model, and a voxel wise ridge regression model is learned to predict the fMRI activations for that TR, from the concatenation of the [CLS] and mean pooled output tokens.
For training and testing, we use data from season 4 of Friends which was unseen during brain-tuning, using 8298 TRs to train and 2630 to test.

\textbf{Normalized Brain Alignment.} Following \cite{moussa2025improvingsemanticunderstandingspeech, oota2025multimodalbrainencodingmodels} prediction performance of this encoding model on the test data is computed by voxel-wise Pearson correlation between the predicted fMRI activations, and the corresponding real brain responses. To account for different levels of noise between voxels, this voxel-wise correlation is then divided by the voxel-wise cross subject prediction accuracy, and averaged across all voxels in each ROI to provide a standardized measure for alignment between the model and different ROIs. We report normalized brain-alignment scores for two subdivisions of the STS--the anterior STS (aSTS), and posterior STS (pSTS), as well as to two adjacent ROIs in the lateral stream (LOC, EBA). For each subject, we visualize the difference in normalized alignment between our brain-tuned models and pretrained (brain-tuned - pretrained) over the entire brain surface. Following \cite{moussa2025improvingsemanticunderstandingspeech}, to test whether the brain-tuned models have significantly improved alignment to an ROI compared to our baselines, for each baseline we perform a wilcoxon signed rank test over the alignment of our brain-tuned models compared to the baseline models' alignment. We indicate significant differences (p < 0.05) with an asterisk \significantAsterisk.

\subsection{Downstream Evaluation}

\textbf{Sarcasm Detection.} We first evaluate our brain-tuned and baseline models on MUStARD \cite{mustard}, an audio-video sarcasm detection database consisting of clips from various sitcoms. Because our models are brain-tuned to stimuli from a sitcom, this measures how our model's performance changes on a social cognition task with stimuli similar to the stimulus seen during brain-tuning. Each clip contains an utterance, accompanied by conversational context, and is labeled for the sarcasm of the utterance. Because some MUStARD clips are from \Friends, we train and test our classifier separately on both the full dataset, and a subset of the dataset with all Friends clips removed. Due to the small size of the dataset, models are evaluated on their mean performance across 10-fold cross validation.

\textbf{Sentiment and Emotion Detection.} To probe social cognition on our baseline and brain-tuned models' in a task markedly different from the \Friends training data, we evaluate on CMU-MOSEI sentiment and emotion prediction \cite{cmu_mosei}, a dataset containing clips of people speaking into the camera from YouTube, and manually labeled for scalar sentiment, and the presence of each of six emotions (happy, sad, anger, surprise, disgust, fear). We use the original 15,288/4,830 train-test split provided by the original TVLT paper \cite{tang2022tvlt}.

\textbf{Evaluation Protocol.} For both tasks, we a train a linear binary classifier on a concatenation of the [CLS] token and mean pooled tokens from the last layer. Since we brain-tune models through mean pooled tokens, but pretrained \TVLT typically probes its [CLS] token for classification tasks, we concatenate both to fairly compare to baselines. We report A2 accuracy and F1 score for our binary classification tasks (sentiment, sarcasm), and weighted A2 accuracy and F1 score for emotion, averaged across n=6 for our brain-tuned models. We use a one-sided one sample t-test over the change in performance of our n=6 subject models compared to each baseline to test for significance, indicating significant improvements (p < 0.05) with an asterisk \significantAsterisk, and highly significant improvements (p < 0.01) with a double asterisk \highlySignificantAsterisk in our graphs. Error bars report SEM across n=6 brain-tuned models.

\section{Results}

\textbf{Brain Alignment Results.} 
We plot the change in alignment compared to the pretrained model (brain-tuned - pretrained) over the entire cortex for subject 03 in \cref{fig:brain_tuning_single_subject}, with other subjects plotted in
\cref{fig:all_subjects_alignment_differences}. Using cross-subject prediction accuracy underestimates the true noise ceiling, as some biological signal that varies between subjects is treated as noise. This leads to some normalized brain alignment scores above 1.0 for baseline and brain-tuned models, but their relative performance is unaffected by this scaling. 
Compared to both pretrained and stimulus tuned baselines, our n=6 brain-tuned models show significant improvements to brain alignment across various lateral stream ROIs (\cref{fig:brain_tuning_single_subject}). We report significantly increased alignment (p<0.05) to both subregions (aSTS, pSTS) of the STS (tuning target), and to one of two neighboring ROIs in the lateral stream (LOC). 
We observe no significant changes in alignment between our pretrained and stimulus-tuned baselines, confirming that increased brain-alignment in our brain-tuned models is not merely due to stimulus exposure.

\begin{figure}
    \centering
    \begin{tabular}{c c}
        a) \includegraphics[width=0.4\linewidth]{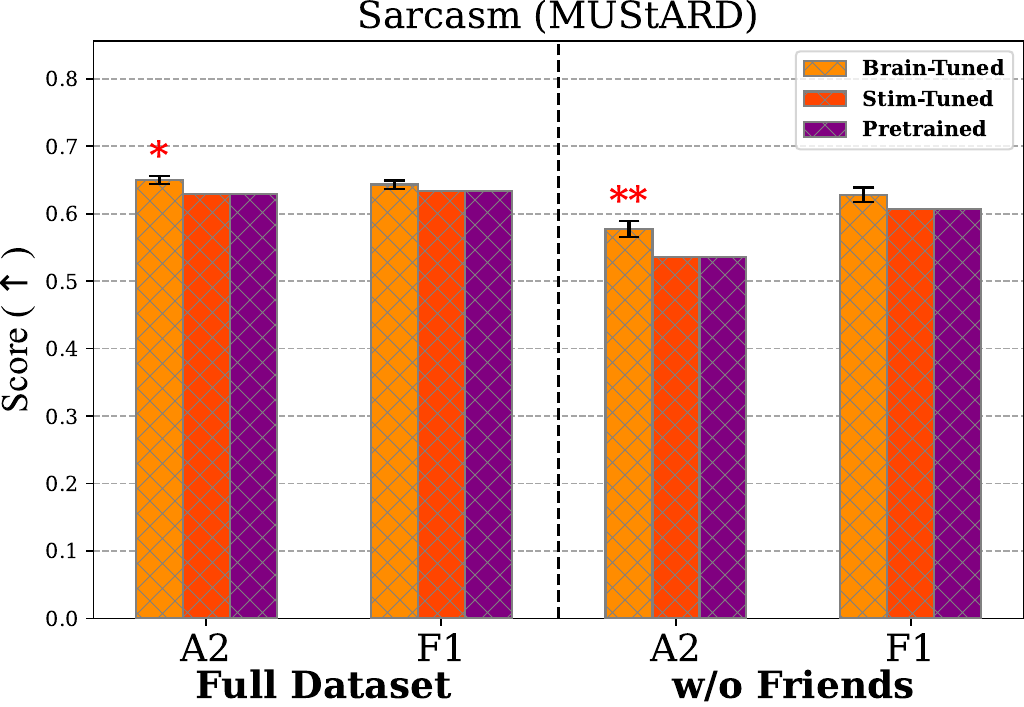} &
        b) \includegraphics[width=0.4\linewidth]{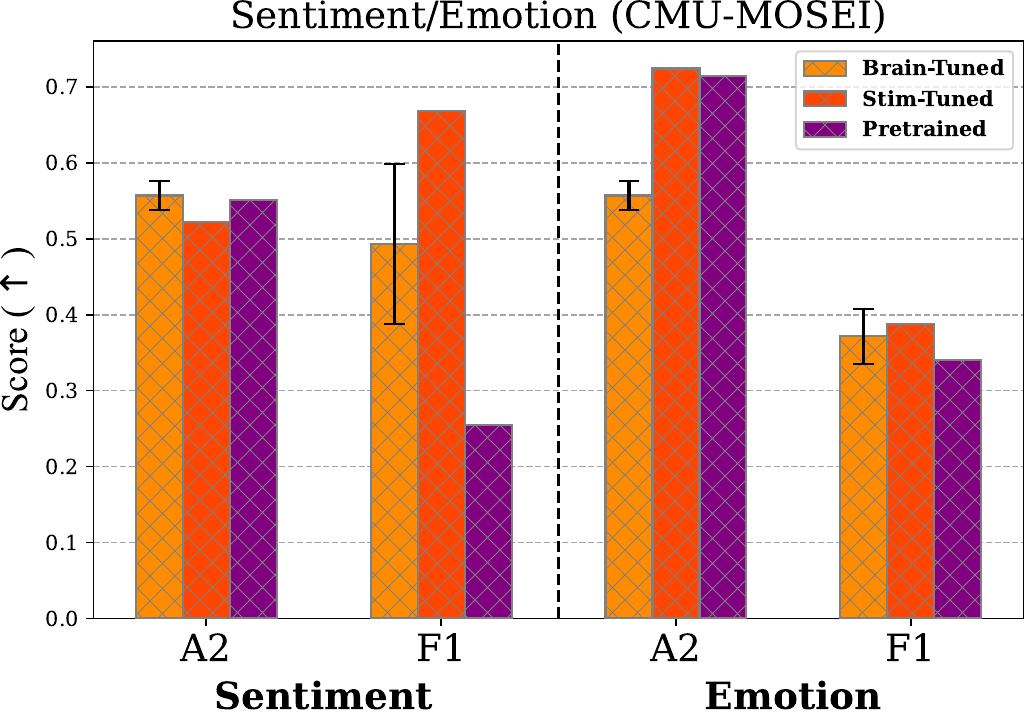} 
    \end{tabular}
    \caption{Brain-tuned and baseline performance on downstream social perception benchmarks. We find significant improvements on \Mustard A2 scores both including \Friends clips ($p<0.05$) and omitting them ($p<0.01$). 
    }\label{fig:downstream_tasks_bar}
\end{figure}

\textbf{Downstream Tasks Results.} Our brain-tuned models significantly outperform baselines on both the full MUStARD sarcasm detection dataset (p<0.05), as well as the dataset after removing all \Friends clips (p<0.01) (\cref{fig:downstream_tasks_bar}a). In contrast, we observe no improvements or decreased performance on the sentiment and emotion prediction task (CMU-MOSEI). These results suggest our model improves performance on a social cognition task similar to the training stimulus (MUStARD), but that these increases do not generalize to a markedly different context (CMU-MOSEI). In the appendix, we break down our emotion classification results by individual emotion.

\section{Conclusion}\label{conclusion}
Our findings demonstrate that brain-tuning a multimodal audio-video model to a social cognition region (STS) not only increases alignment to the target area but also extends improved alignment to an adjacent lateral stream ROI. This increased alignment is accompanied by significant gains on a related social cognition task when the evaluation context resembles the training stimulus, sarcasm detection in sitcoms. However, these gains do not generalize to sentiment and emotion prediction in markedly different contexts, suggesting a limitation in the transferability of brain-tuning effects to contexts unseen during training. While our study was limited to a single model and a small number of evaluations, the results serve as a proof of concept for targeted brain-tuning as a means to enhance both brain alignment and task performance in relevant domains. We suggest future researchers experiment with larger LLM based multi-modal architectures, as well as more diverse evaluation and training datasets.

\section{Funding and Acknowledgments}

Funded by the Deutsche Forschungsgemeinschaft (DFG, German Research Foundation) -- GRK 2853/1 “Neuroexplicit Models of Language, Vision, and Action” - project number 471607914.

The Courtois project on neural modelling was made possible by a generous donation from the Courtois foundation, administered by the Fondation Institut Gériatrie Montréal at CIUSSS du Centre-Sud-de-l’île-de-Montréal and University of Montreal. The Courtois NeuroMod team is based at “Centre de Recherche de l’Institut Universitaire de Gériatrie de Montréal”, with several other institutions involved. See the cneuromod documentation for an up-to-date list of contributors (\url{https://docs.cneuromod.ca}).

The authors gratefully acknowledge support from the DAAD RISE (Research Internships in Science and Engineering) program. The authors greatfully acknowledge support from MPI for Software Systems and MPI for Informatics. The authors thank Omer Moussa, Dota Dong, and Subba Oota Reddy for their helpful advice and guidance during the development of this study.

{
    \small
    \bibliographystyle{ieeenat_fullname}
    \bibliography{main}

@String(CVPR= {IEEE Conf. Comput. Vis. Pattern Recog.})

@String(CVPR  = {CVPR})

@article{bae2016pt735,
  title={PT735. Multisensory integration of social interaction},
  author={Bae, Dahye and Kwon, Jun Soo},
  journal={International Journal of Neuropsychopharmacology},
  volume={19},
  number={Suppl 1},
  pages={67},
  year={2016}
}

@article{benetti2023multimodal,
  title={Multimodal processing in face-to-face interactions: A bridging link between psycholinguistics and sensory neuroscience},
  author={Benetti, Stefania and Ferrari, Ambra and Pavani, Francesco},
  journal={Frontiers in Human Neuroscience},
  volume={17},
  pages={1108354},
  year={2023},
  publisher={Frontiers Media SA}
}

@article{bolotta2022social,
  title={Social neuro AI: Social interaction as the “dark matter” of AI},
  author={Bolotta, Samuele and Dumas, Guillaume},
  journal={Frontiers in computer science},
  volume={4},
  pages={846440},
  year={2022},
  publisher={Frontiers Media SA}
}

@article{campanella2007integrating,
  title={Integrating face and voice in person perception},
  author={Campanella, Salvatore and Belin, Pascal},
  journal={Trends in cognitive sciences},
  volume={11},
  number={12},
  pages={535--543},
  year={2007},
  publisher={Elsevier}
}

@misc{dascoli2025tribetrimodalbrainencoder,
      title={TRIBE: TRImodal Brain Encoder for whole-brain fMRI response prediction}, 
      author={Stéphane d'Ascoli and Jérémy Rapin and Yohann Benchetrit and Hubert Banville and Jean-Rémi King},
      year={2025},
      eprint={2507.22229},
      archivePrefix={arXiv},
      primaryClass={cs.LG},
      url={https://arxiv.org/abs/2507.22229}, 
}

@article{hein2008superior,
  title={Superior Temporal Sulcus—It's My Area: Or Is It?},
  author={Hein, Grit and Knight, Robert T},
  journal={Journal of Cognitive Neuroscience},
  volume={20},
  number={12},
  pages={2125--2136},
  year={2008},
  publisher={MIT Press}
}

@article{pitcher2021lateralevidence,
  title={Evidence for a third visual pathway specialized for social perception},
  author={Pitcher, David and Ungerleider, Leslie G},
  journal={Trends in cognitive sciences},
  volume={25},
  number={2},
  pages={100--110},
  year={2021},
  publisher={Elsevier}
}

@article{walbrin2018neural,
  title={Neural responses to visually observed social interactions},
  author={Walbrin, Jon and Downing, Paul and Koldewyn, Kami},
  journal={Neuropsychologia},
  volume={112},
  pages={31--39},
  year={2018},
  publisher={Elsevier}
}

@article{isik2017social,
author = {Leyla Isik  and Kami Koldewyn  and David Beeler  and Nancy Kanwisher },
title = {Perceiving social interactions in the posterior superior temporal sulcus},
journal = {Proceedings of the National Academy of Sciences},
volume = {114},
number = {43},
pages = {E9145-E9152},
year = {2017},
doi = {10.1073/pnas.1714471114},
URL = {https://www.pnas.org/doi/abs/10.1073/pnas.1714471114},
eprint = {https://www.pnas.org/doi/pdf/10.1073/pnas.1714471114}}

@article{negi2025brain,
  title={Brain-Informed Fine-Tuning for Improved Multilingual Understanding in Language Models},
  author={Negi, Anuja and Oota, Subba Reddy and Gupta, Manish and Deniz, Fatma},
  journal={bioRxiv},
  pages={2025--07},
  year={2025},
  publisher={Cold Spring Harbor Laboratory}
}

@article{allison2000social,
  title={Social perception from visual cues: role of the STS region},
  author={Allison, Truett and Puce, Aina and McCarthy, Gregory},
  journal={Trends in cognitive sciences},
  volume={4},
  number={7},
  pages={267--278},
  year={2000},
  publisher={Elsevier}
}

@article{masson2021functional,
  title={Functional selectivity for social interaction perception in the human superior temporal sulcus during natural viewing},
  author={Masson, Haemy Lee and Isik, Leyla},
  journal={NeuroImage},
  volume={245},
  pages={118741},
  year={2021},
  publisher={Elsevier}
}

@article{deen2015functional,
  title={Functional Organization of Social Perception and Cognition in the Superior Temporal Sulcus},
  author={Deen, Ben and Koldewyn, Kami and Kanwisher, Nancy and Saxe, Rebecca},
  journal={Cerebral cortex},
  volume={25},
  number={11},
  pages={4596--4609},
  year={2015},
  publisher={Oxford University Press}
}

@article{antonello2023scaling,
  title={Scaling laws for language encoding models in fMRI},
  author={Antonello, Richard and Vaidya, Aditya and Huth, Alexander},
  journal={Advances in Neural Information Processing Systems},
  volume={36},
  pages={21895--21907},
  year={2023}
}

@inproceedings{boyle2020courtois,
  title={The courtois project on neuronal modelling-first data release},
  author={Boyle, Julie A and Pinsard, Basile and Boukhdhir, Amal and Belleville, Sylvie and Brambatti, Simona and Chen, Jeni and Cohen-Adad, Julien and Cyr, Andr{\'e} and Rainville, Pierre and Bellec, Pierre and others},
  year={2020},
  booktitle = {26th Annual Meeting of the Organization for Human Brain Mapping (OHBM)},
  publisher={Organization for Human Brain Mapping (OHBM)}
}

@misc{hu2024minicpmunveilingpotentialsmall,
      title={MiniCPM: Unveiling the Potential of Small Language Models with Scalable Training Strategies}, 
      author={Shengding Hu and Yuge Tu and Xu Han and Chaoqun He and Ganqu Cui and Xiang Long and Zhi Zheng and Yewei Fang and Yuxiang Huang and Weilin Zhao and Xinrong Zhang and Zheng Leng Thai and Kaihuo Zhang and Chongyi Wang and Yuan Yao and Chenyang Zhao and Jie Zhou and Jie Cai and Zhongwu Zhai and Ning Ding and Chao Jia and Guoyang Zeng and Dahai Li and Zhiyuan Liu and Maosong Sun},
      year={2024},
      eprint={2404.06395},
      archivePrefix={arXiv},
      primaryClass={cs.CL},
      url={https://arxiv.org/abs/2404.06395}, 
}

@misc{liu2025olapushingfrontiersomnimodal,
      title={Ola: Pushing the Frontiers of Omni-Modal Language Model}, 
      author={Zuyan Liu and Yuhao Dong and Jiahui Wang and Ziwei Liu and Winston Hu and Jiwen Lu and Yongming Rao},
      year={2025},
      eprint={2502.04328},
      archivePrefix={arXiv},
      primaryClass={cs.CV},
      url={https://arxiv.org/abs/2502.04328}, 
}

@misc{li2025baichuanomni15technicalreport,
      title={Baichuan-Omni-1.5 Technical Report}, 
      author={Yadong Li and others},
      year={2025},
      eprint={2501.15368},
      archivePrefix={arXiv},
      primaryClass={cs.CL},
      url={https://arxiv.org/abs/2501.15368}, 
}

@misc{shu2023audiovisualllmvideounderstanding,
      title={Audio-Visual LLM for Video Understanding}, 
      author={Fangxun Shu and Lei Zhang and Hao Jiang and Cihang Xie},
      year={2023},
      eprint={2312.06720},
      archivePrefix={arXiv},
      primaryClass={cs.CV},
      url={https://arxiv.org/abs/2312.06720}, 
}

@misc{caffagni2024revolutionmultimodallargelanguage,
      title={The Revolution of Multimodal Large Language Models: A Survey}, 
      author={Davide Caffagni and Federico Cocchi and Luca Barsellotti and Nicholas Moratelli and Sara Sarto and Lorenzo Baraldi and Lorenzo Baraldi and Marcella Cornia and Rita Cucchiara},
      year={2024},
      eprint={2402.12451},
      archivePrefix={arXiv},
      primaryClass={cs.CV},
      url={https://arxiv.org/abs/2402.12451}, 
}

@article{multimodal_learning_survey,
    author = {Yuan, Yuan and Li, Zhaojian and Zhao, Bin},
    year = {2025},
    month = {01},
    pages = {},
    title = {A Survey of Multimodal Learning: Methods, Applications, and Future},
    volume = {57},
    journal = {ACM Computing Surveys},
    doi = {10.1145/3713070}
    }

@misc{lu2022multimodalfoundationmodelsbetter,
      title={Multimodal foundation models are better simulators of the human brain}, 
      author={Haoyu Lu and Qiongyi Zhou and Nanyi Fei and Zhiwu Lu and Mingyu Ding and Jingyuan Wen and Changde Du and Xin Zhao and Hao Sun and Huiguang He and Ji-Rong Wen},
      year={2022},
      eprint={2208.08263},
      archivePrefix={arXiv},
      primaryClass={cs.NE},
      url={https://arxiv.org/abs/2208.08263}, 
}

@misc{fu2025vitaopensourceinteractiveomni,
      title={VITA: Towards Open-Source Interactive Omni Multimodal LLM}, 
      author={Chaoyou Fu and Haojia Lin and Zuwei Long and Yunhang Shen and Yuhang Dai and Meng Zhao and Yi-Fan Zhang and Shaoqi Dong and Yangze Li and Xiong Wang and Haoyu Cao and Di Yin and Long Ma and Xiawu Zheng and Rongrong Ji and Yunsheng Wu and Ran He and Caifeng Shan and Xing Sun},
      year={2025},
      eprint={2408.05211},
      archivePrefix={arXiv},
      primaryClass={cs.CV},
      url={https://arxiv.org/abs/2408.05211}, 
}

@misc{dong2023visionlanguageintegrationmultimodalvideo,
      title={Vision-Language Integration in Multimodal Video Transformers (Partially) Aligns with the Brain}, 
      author={Dota Tianai Dong and Mariya Toneva},
      year={2023},
      eprint={2311.07766},
      archivePrefix={arXiv},
      primaryClass={cs.CV},
      url={https://arxiv.org/abs/2311.07766}, 
}

@misc{moussa2025improvingsemanticunderstandingspeech,
      title={Improving Semantic Understanding in Speech Language Models via Brain-tuning}, 
      author={Omer Moussa and Dietrich Klakow and Mariya Toneva},
      year={2025},
      eprint={2410.09230},
      archivePrefix={arXiv},
      primaryClass={cs.CL},
      url={https://arxiv.org/abs/2410.09230}, 
}

@ARTICLE{Porter2023-qw,
  title     = "Emotion detection and its influence on popularity in a social
               network-based on the American {TV} series Friends",
  author    = "Porter, Ilana and Galam, Bar and Ramon-Gonen, Roni",
  journal   = "Soc. Netw. Anal. Min.",
  publisher = "Springer Science and Business Media LLC",
  volume    =  13,
  number    =  1,
  month     =  sep,
  year      =  2023,
  copyright = "https://www.springernature.com/gp/researchers/text-and-data-mining",
  language  = "en"
}

@inproceedings{
garcia2025modeling,
title={Modeling dynamic social vision highlights gaps between deep learning and humans},
author={Kathy Garcia and Emalie McMahon and Colin Conwell and Michael Bonner and Leyla Isik},
booktitle={The Thirteenth International Conference on Learning Representations},
year={2025},
url={https://openreview.net/forum?id=wAXsx2MYgV}
}

@misc{oota2025multimodalbrainencodingmodels,
      title={Multi-modal brain encoding models for multi-modal stimuli}, 
      author={Subba Reddy Oota and Khushbu Pahwa and Mounika Marreddy and Maneesh Singh and Manish Gupta and Bapi S. Raju},
      year={2025},
      eprint={2505.20027},
      archivePrefix={arXiv},
      primaryClass={q-bio.NC},
      url={https://arxiv.org/abs/2505.20027}, 
}

@inproceedings{nakagi-etal-2024-unveiling,
    title = "Unveiling Multi-level and Multi-modal Semantic Representations in the Human Brain using Large Language Models",
    author = "Nakagi, Yuko  and
      Matsuyama, Takuya  and
      Koide-Majima, Naoko  and
      Yamaguchi, Hiroto Q.  and
      Kubo, Rieko  and
      Nishimoto, Shinji  and
      Takagi, Yu",
    editor = "Al-Onaizan, Yaser  and
      Bansal, Mohit  and
      Chen, Yun-Nung",
    booktitle = "Proceedings of the 2024 Conference on Empirical Methods in Natural Language Processing",
    month = nov,
    year = "2024",
    address = "Miami, Florida, USA",
    publisher = "Association for Computational Linguistics",
    url = "https://aclanthology.org/2024.emnlp-main.1133/",
    doi = "10.18653/v1/2024.emnlp-main.1133",
    pages = "20313--20338",
}

@misc{sun2024videosalmonnspeechenhancedaudiovisuallarge,
      title={video-SALMONN: Speech-Enhanced Audio-Visual Large Language Models}, 
      author={Guangzhi Sun and Wenyi Yu and Changli Tang and Xianzhao Chen and Tian Tan and Wei Li and Lu Lu and Zejun Ma and Yuxuan Wang and Chao Zhang},
      year={2024},
      eprint={2406.15704},
      archivePrefix={arXiv},
      primaryClass={cs.CV},
      url={https://arxiv.org/abs/2406.15704}, 
}

@misc{lyu2023macawllmmultimodallanguagemodeling,
      title={Macaw-LLM: Multi-Modal Language Modeling with Image, Audio, Video, and Text Integration}, 
      author={Chenyang Lyu and Minghao Wu and Longyue Wang and Xinting Huang and Bingshuai Liu and Zefeng Du and Shuming Shi and Zhaopeng Tu},
      year={2023},
      eprint={2306.09093},
      archivePrefix={arXiv},
      primaryClass={cs.CL},
      url={https://arxiv.org/abs/2306.09093}, 
}

@article{schrimpf2021neural,
  title={The neural architecture of language: Integrative modeling converges on predictive processing},
  author={Schrimpf, Martin and Blank, Idan Asher and Tuckute, Greta and Kauf, Carina and Hosseini, Eghbal A and Kanwisher, Nancy and Tenenbaum, Joshua B and Fedorenko, Evelina},
  journal={Proceedings of the National Academy of Sciences},
  volume={118},
  number={45},
  pages={e2105646118},
  year={2021},
  publisher={National Academy of Sciences}
}

@inproceedings{mustard,
    title = "Towards Multimodal Sarcasm Detection (An {\_}{O}bviously{\_} Perfect Paper)",
    author = "Castro, Santiago  and
      Hazarika, Devamanyu  and
      P{\'e}rez-Rosas, Ver{\'o}nica  and
      Zimmermann, Roger  and
      Mihalcea, Rada  and
      Poria, Soujanya",
    editor = "Korhonen, Anna  and
      Traum, David  and
      M{\`a}rquez, Llu{\'i}s",
    booktitle = "Proceedings of the 57th Annual Meeting of the Association for Computational Linguistics",
    month = jul,
    year = "2019",
    address = "Florence, Italy",
    publisher = "Association for Computational Linguistics",
    url = "https://aclanthology.org/P19-1455/",
    doi = "10.18653/v1/P19-1455",
    pages = "4619--4629"
}

@inproceedings{cmu_mosei,
    title = "Multimodal Language Analysis in the Wild: {CMU}-{MOSEI} Dataset and Interpretable Dynamic Fusion Graph",
    author = "Bagher Zadeh, AmirAli  and
      Liang, Paul Pu  and
      Poria, Soujanya  and
      Cambria, Erik  and
      Morency, Louis-Philippe",
    editor = "Gurevych, Iryna  and
      Miyao, Yusuke",
    booktitle = "Proceedings of the 56th Annual Meeting of the Association for Computational Linguistics (Volume 1: Long Papers)",
    month = jul,
    year = "2018",
    address = "Melbourne, Australia",
    publisher = "Association for Computational Linguistics",
    url = "https://aclanthology.org/P18-1208/",
    doi = "10.18653/v1/P18-1208",
    pages = "2236--2246"
}

@misc{zellers2022merlotreserveneuralscript,
      title={MERLOT Reserve: Neural Script Knowledge through Vision and Language and Sound}, 
      author={Rowan Zellers and Jiasen Lu and Ximing Lu and Youngjae Yu and Yanpeng Zhao and Mohammadreza Salehi and Aditya Kusupati and Jack Hessel and Ali Farhadi and Yejin Choi},
      year={2022},
      eprint={2201.02639},
      archivePrefix={arXiv},
      primaryClass={cs.CV},
      url={https://arxiv.org/abs/2201.02639}, 
}

@misc{xu2025qwen25omnitechnicalreport,
      title={Qwen2.5-Omni Technical Report}, 
      author={Jin Xu and Zhifang Guo and Jinzheng He and Hangrui Hu and Ting He and Shuai Bai and Keqin Chen and Jialin Wang and Yang Fan and Kai Dang and Bin Zhang and Xiong Wang and Yunfei Chu and Junyang Lin},
      year={2025},
      eprint={2503.20215},
      archivePrefix={arXiv},
      primaryClass={cs.CL},
      url={https://arxiv.org/abs/2503.20215}, 
}

@misc{cheng2024videollama2advancingspatialtemporal,
      title={VideoLLaMA 2: Advancing Spatial-Temporal Modeling and Audio Understanding in Video-LLMs}, 
      author={Zesen Cheng and Sicong Leng and Hang Zhang and Yifei Xin and Xin Li and Guanzheng Chen and Yongxin Zhu and Wenqi Zhang and Ziyang Luo and Deli Zhao and Lidong Bing},
      year={2024},
      eprint={2406.07476},
      archivePrefix={arXiv},
      primaryClass={cs.CV},
      url={https://arxiv.org/abs/2406.07476}, 
}

@inproceedings{girdhar2023imagebind,
  title={ImageBind: One Embedding Space To Bind Them All},
  author={Girdhar, Rohit and El-Nouby, Alaaeldin and Liu, Zhuang
and Singh, Mannat and Alwala, Kalyan Vasudev and Joulin, Armand and Misra, Ishan},
  booktitle={CVPR},
  year={2023}
}

@inproceedings{tang2022tvlt,
  title     = {TVLT: Textless Vision-Language Transformer},
  author    = {Zineng Tang and Jaemin Cho and Yixin Nie and Mohit Bansal},
  booktitle = {NeurIPS},
  year      = {2022}
}

@article{tuningfreteault,
    author = {Freteault, Maëlle and Le Clei, Maximilien and Tetrel, Loic and Bellec, Lune and Farrugia, Nicolas},
    title = {Alignment of auditory artificial networks with massive individual fMRI brain data leads to generalisable improvements in brain encoding and downstream tasks},
    journal = {Imaging Neuroscience},
    volume = {3},
    pages = {imag\_a\_00525},
    year = {2025},
    month = {04},
    issn = {2837-6056},
    doi = {10.1162/imag_a_00525},
    url = {https://doi.org/10.1162/imag\_a\_00525},
    eprint = {https://direct.mit.edu/imag/article-pdf/doi/10.1162/imag\_a\_00525/2508588/imag\_a\_00525.pdf},
}

@article{oota2025correlating,
  title={Correlating instruction-tuning (in multimodal models) with vision-language processing (in the brain)},
  author={Oota, Subba Reddy and Jindal, Akshett and Mondal, Ishani and Pahwa, Khushbu and Namburi, Satya Sai Srinath and Shrivastava, Manish and Singh, Maneesh and Raju, Bapi S and Gupta, Manish},
  journal={arXiv preprint arXiv:2505.20029},
  year={2025}
}

@inbook{voss_2016,
    author = {Voss, Michelle},
    year = {2016},
    month = {12},
    pages = {187-209},
    publisher = {Elsevier Academic Press},
    title = {The Chronic Exercise–Cognition Interaction},
    isbn = {9780128007785},
    doi = {10.1016/B978-0-12-800778-5.00009-8}
}
}

\appendix

\clearpage
\setcounter{page}{1}

\section{Appendix}

This appendix contains the following sections: 

In \cref{appendix:participants}, we provide additional information on the subjects used in the fMRI data collection. 

In \cref{appendix:noise_ceilings}, we visualize the effect of the accuracy threshold on the number of voxels we can brain-tune on.

In \cref{subsec:all_brain_alignments}, we display the changes in brain alignment for all subjects.

In \cref{subsec:cmu_mosei_all_emotions}, we expand on the CMU MOSEI evaluation results across all emotions.

In \cref{subsec:sts_mask_visualization}, we show the STS region mask.

In \cref{subsec:broader_impacts}, we discuss the potential broader impacts of our work.

In \cref{subsec:data_and_code_availability}, we provide information on how to access our code and relevant data.

In \cref{subsec:licenses}, we include all licensing information.

\subsection{Participants}\label{appendix:participants}
Six healthy participants (aged 31 to 47 years at the time
of recruitment in 2018), three women (sub-03, sub-04,
and sub-06) and three men (sub-01, sub-02, and sub-05)
were recruited to participate in the Courtois Neuromod
Project for at least 5 years. All subjects provided informed
consent to participate in this study, which was approved
by the ethics review board of the “CIUSS du centre-sud-
de- l’île- de- Montréal” (under number CER VN 18-19-22).
Three of the participants reported being native franco-
phone speakers (sub- 01, sub-02, and sub-04), one as
being a native anglophone (sub-06), and two as bilingual
native speakers (sub-03 and sub-05). All participants
reported the right hand as being their dominant hand and
reported being in good general health. Exclusion criteria
included visual or auditory impairments that would prevent participants from seeing and/or hearing stimuli in the
scanner and major psychiatric or neurological problems.
Standard exclusion criteria for MRI and MEG were also
applied. Lastly, given that all stimuli and instructions are
presented in English, all participants had to report having
an advanced comprehension of the English language for
inclusion. The above boilerplate text is taken from the
cNeuroMod documentation \cite{boyle2020courtois}, with the
express intention that users should copy and paste this
text into their manuscripts unchanged. It was released by
the Courtois NeuroMod team under the CC0 license. 

\subsection{Cross Subject Prediction Accuracy Calculation}\label{appendix:noise_ceilings}
\begin{figure}
    \centering
    \includegraphics[width=0.9\linewidth]{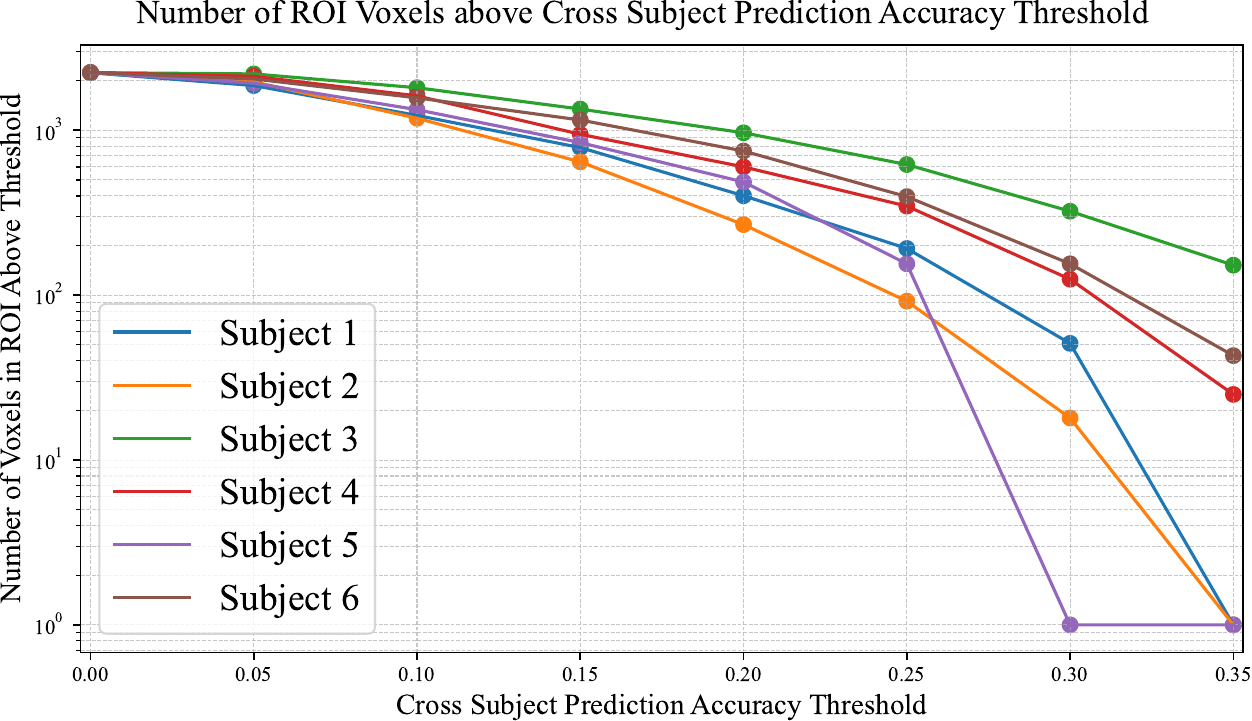}
    \caption{A subject (Subject $5$) has no voxels in the STS above a cross subject prediction accuracy threshold of 0.25, and thus we cannot perform brain-tuning.}
    \label{fig:noise_ceilings_over_subjects}
\end{figure}

We follow recent studies \cite{oota2025multimodalbrainencodingmodels}, \cite{dong2023visionlanguageintegrationmultimodalvideo} in adapting \cite{schrimpf2021neural}'s method to estimate cross-subject prediction accuracy for each voxel. For each subject, we generate all possible subsets of the remaining 5 subjects, and for each subset we use a voxel-wise encoding model (see Sec. 5) to predict one participant’s
response from the others. As in previous studies \cite{dong2023visionlanguageintegrationmultimodalvideo, oota2025multimodalbrainencodingmodels}, the final value is calculated as an average at the group level. These cross-subject encoding models are trained using nine episodes (7700 TRs) from the first season of Friends, and tested on three other episodes from the same season (2872 TRs). 
In \cref{fig:noise_ceilings_over_subjects}, we display the number of viable voxels based on the cross subject prediction accuracy threshold. We observe that Subject-05 has no remaining voxels above $0.25$, and thus deem this as our cut-off.

\subsection{Differences in Normalized Brain Alignment For all Subjects}\label{subsec:all_brain_alignments}

\begin{figure}
    \centering
    \begin{tabular}{cc}
        \multicolumn{2}{c}{Difference in Brain Alignment(Brain-tuned - Pretrained)} \\
        \includegraphics[width=\allSubjectsBrainWidth]{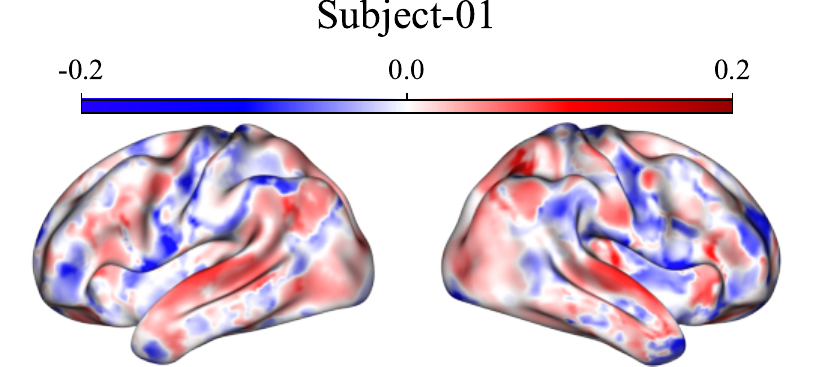} & 
        \includegraphics[width=\allSubjectsBrainWidth]{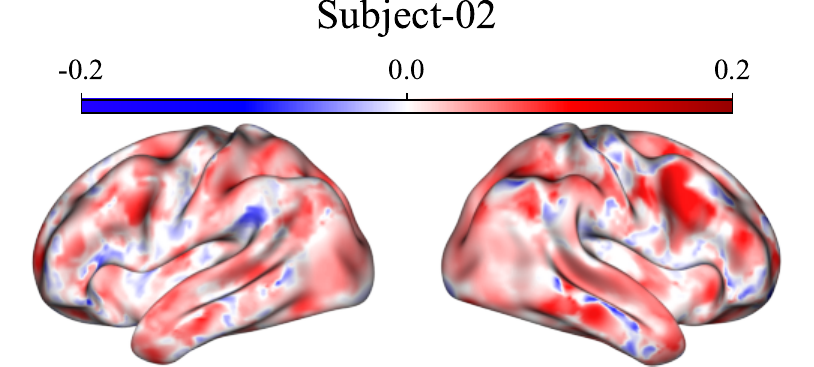} \\ 
        \includegraphics[width=\allSubjectsBrainWidth]{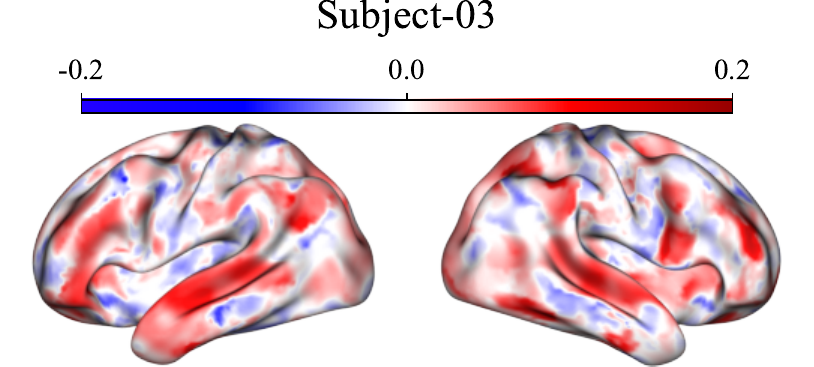} &
        \includegraphics[width=\allSubjectsBrainWidth]{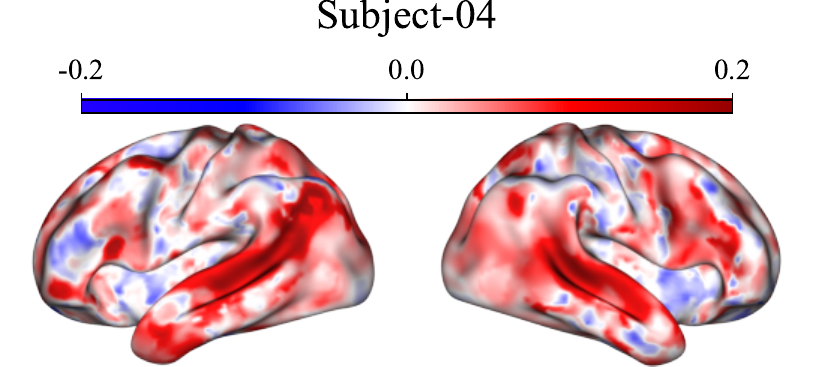} \\ 
        \includegraphics[width=\allSubjectsBrainWidth]{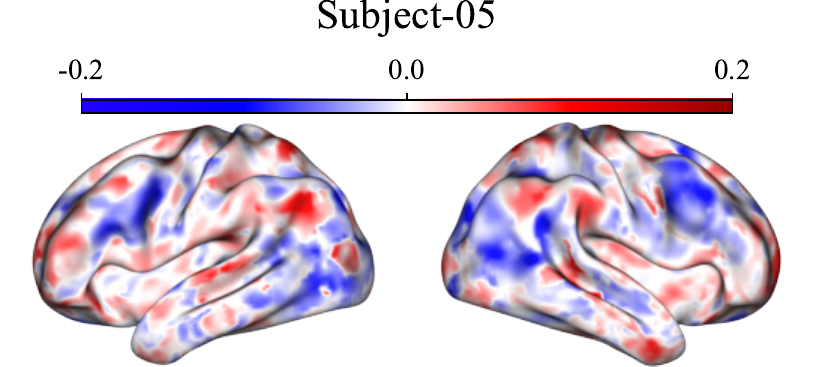} & 
        \includegraphics[width=\allSubjectsBrainWidth]{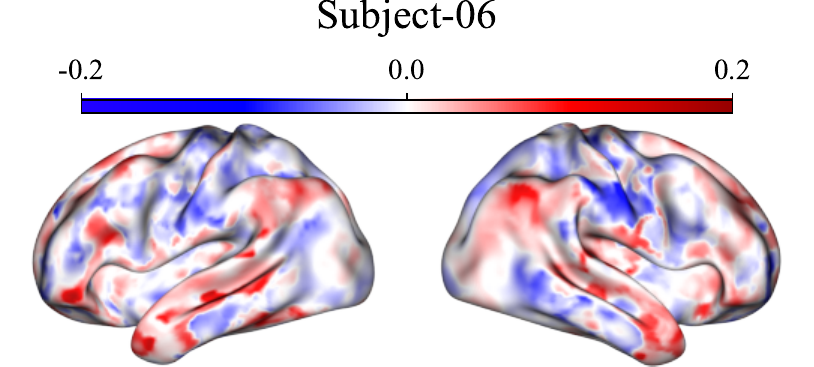} \\ 
    \end{tabular}
    \caption{Differences in Normalized Brain Alignment before and after brain-tuning.}
    \label{fig:all_subjects_alignment_differences}
\end{figure}

In \cref{fig:all_subjects_alignment_differences}, we display the differences in normalized brain alignment for all subjects. Most subject models show improved alignment in and around the STS, but these improvements do not consistently extend to other regions.

\subsection{CSU MOSEI Complete Emotion}\label{subsec:cmu_mosei_all_emotions}

\begin{figure}
    \centering
    \includegraphics[width=0.8\linewidth]{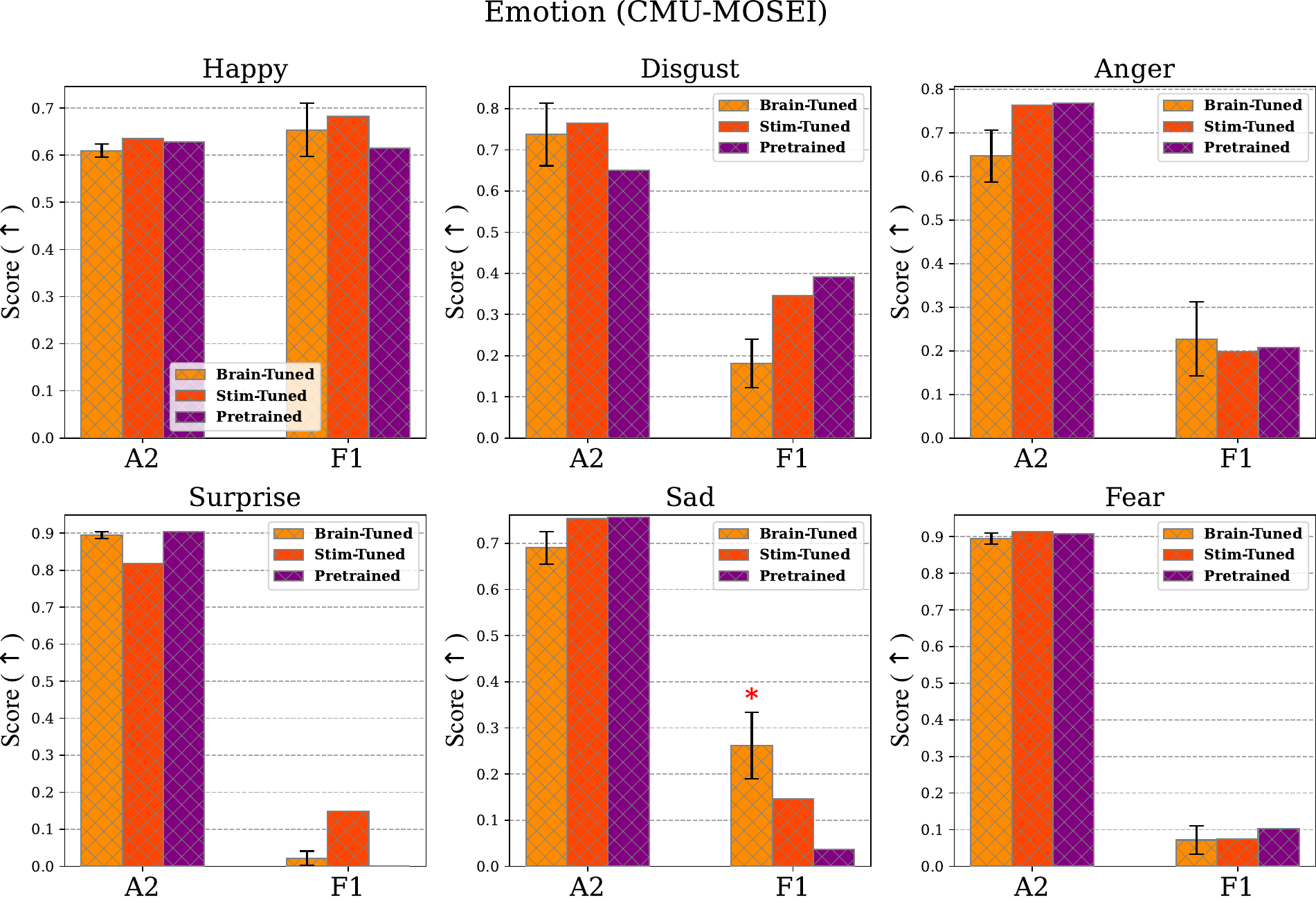}
    \caption{A breakdown of the performance of the model across each emotion aggregated in the CSU MOSEI evaluation (See \cref{fig:downstream_tasks_bar}, rightmost chart). }
    \label{fig:all_emotions}
\end{figure}

In \cref{fig:all_emotions}, we breakdown the performance of the model across each emotion aggregated in the CMU MOSEI evaluation (See \cref{fig:downstream_tasks_bar}, rightmost chart). Notably, sadness is the only emotion with significantly improved F1 score - however, we also observe decreased accuracy (A2). Although sadness occurs in \Friends, it is it not the dominant emotion in the show \cite{Porter2023-qw}. In future work, we hope to investigate this finding further for an explanation.

\subsection{Visualization STS Region}\label{subsec:sts_mask_visualization}
\begin{figure}
    \centering
    \includegraphics[width=0.7\linewidth]{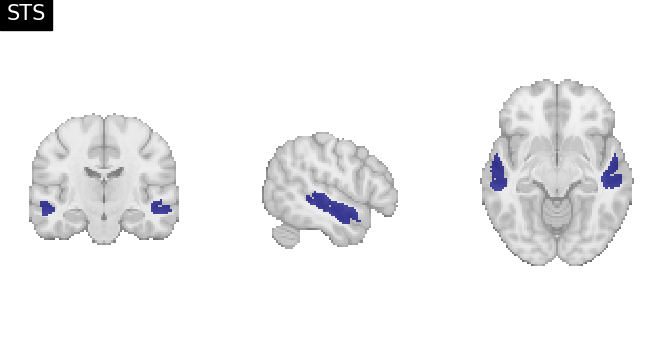}
    \includegraphics[width=0.7\linewidth]{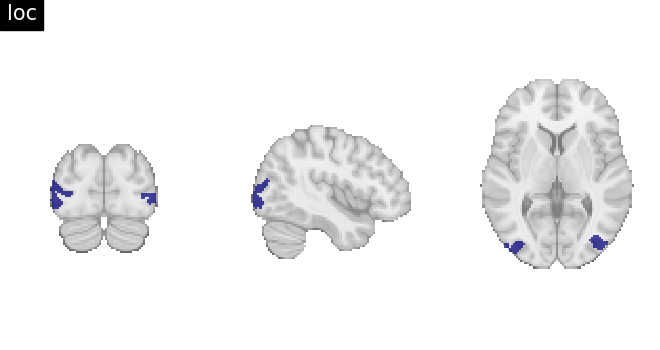}
    \includegraphics[width=0.7\linewidth]{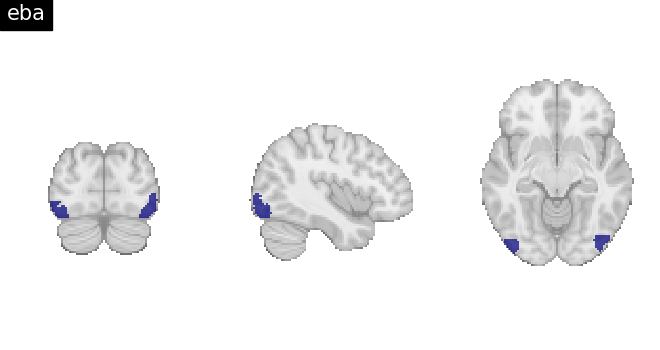}
    \caption{The STS, LOC, and EBA regions from coronal (left),  lateral (middle), and horizontal (right) views.}
    \label{fig:sts_mask}
\end{figure}

 We display the voxel mask of the STS region which we tune our model to in \cref{fig:sts_mask}. 

\subsection{Broader Impacts}\label{subsec:broader_impacts}

Our work can be an initial step towards creating AI models with better understanding of human social cognition using brain activity as a tuning target. This could have positive impacts, such as improving AI-human communication or potential uses in AI-assisted therapy. Further, an AI model which can replicate human social cognition may be a useful in-silico model helping us understand social cognition in humans. On the other hand, this could enhance the abilities of AI for human manipulation. We urge future researchers to consider these pros and cons as they continue investigating this topic.

\subsection{Data and Code Availability}\label{subsec:data_and_code_availability}
The fMRI data used to perform the brain tuning are openly available through registered access at link \url{https://www.cneuromod.ca/access/access/}.

To get our code, and for exact instructions on how to replicate or results, please visit
\url{https://huggingface.co/AnonymousSubmission43/mmbt}, download and unzip all files, and follow the instructions on the README.md in mmbt-anon.

Due to privacy concerns, we do not release model weights or cross subject prediction accuracies, as these are derived from subjects' brain data. 

\subsection{Licenses}\label{subsec:licenses}

\textbf{Method Diagram}: Our method diagram in \cref{fig:brain_tuning_method} includes an audio sound wave, licensed under the public domain. You can find the url here: \url{https://www.pngfind.com/maxpin/bwRwR/}

Models:\\
\begin{itemize}
    \item TVLT: MIT \url{https://github.com/zinengtang/TVLT/blob/main/LICENSE}
\end{itemize}

Packages: \\
\begin{itemize}
    \item nilearn: BSD \url{https://github.com/nilearn/nilearn/blob/main/LICENSE}
    \item matplotlib: BSD \url{https://github.com/nilearn/nilearn/blob/main/LICENSE}
\end{itemize}

Datasets:
\begin{itemize}
    \item CMU-MOSEI: MIT \url{https://github.com/CMU-MultiComp-Lab/CMU-MultimodalSDK?tab=MIT-1-ov-file}
    \item MUSTARD: MIT\url{https://github.com/soujanyaporia/MUStARD/blob/master/LICENSE}
    \item Courtois NeuroMod: CC0 \url{https://creativecommons.org/publicdomain/zero/1.0/legalcode}
\end{itemize}

\newpage
\begin{enumerate}

\item {\bf Claims}
    \item[] Question: Do the main claims made in the abstract and introduction accurately reflect the paper's contributions and scope?
    \item[] Answer: \answerYes{} 
    \item[] Justification: In the abstract and intro, we make claims which are backed up by our experimental results - we claim increases to alignment and improvements to one of two social perception tasks, which accurately reflect the results we find through statistical testing.

\item {\bf Limitations}
    \item[] Question: Does the paper discuss the limitations of the work performed by the authors?
    \item[] Answer: \answerYes{} 
    \item[] Justification: Yes, see conclusion (\cref{conclusion}) for a discussion of the studies limitations (we use single model and a small number of evaluations). 

\item {\bf Theory assumptions and proofs}
    \item[] Question: For each theoretical result, does the paper provide the full set of assumptions and a complete (and correct) proof?
    \item[] Answer: \answerNA{}{} 
    \item[] Justification: We do not include any theoretical results in our paper that require justification.

    \item {\bf Experimental result reproducibility}
    \item[] Question: Does the paper fully disclose all the information needed to reproduce the main experimental results of the paper to the extent that it affects the main claims and/or conclusions of the paper (regardless of whether the code and data are provided or not)?
    \item[] Answer: \answerYes{}{} 
    \item[] Justification: In the methods section (\cref{method}), we fully describe our tuning architecture, the hyper parameters used during training, the specific datasets we use, how we calculate the cross subject prediction accuracies, and make all code publicly available. We do not release any weights or cross subject prediction accuracies derived from fMRI due to privacy concerns, but you can reproduce our experiments following the instructions in our paper and our code. See \cref{subsec:data_and_code_availability}.

\item {\bf Open access to data and code}
    \item[] Question: Does the paper provide open access to the data and code, with sufficient instructions to faithfully reproduce the main experimental results, as described in supplemental material?
    \item[] Answer: \answerYes{} 
    \item[] Justification: We provide open access to our code, with instructions describing how to run the main experiments in the documentation, see \cref{subsec:data_and_code_availability}.

\item {\bf Experimental setting/details}
    \item[] Question: Does the paper specify all the training and test details (e.g., data splits, hyperparameters, how they were chosen, type of optimizer, etc.) necessary to understand the results?
    \item[] Answer: \answerYes{}{} 
    \item[] Justification: In \cref{subsec:braintuning} we provide details on training and testing, including the data splits, optimizer type, the choice of hyperparameters, and how they were chosen. Furthermore, one can view our provided code documentation for these settings, see \cref{subsec:data_and_code_availability}.

\item {\bf Experiment statistical significance}
    \item[] Question: Does the paper report error bars suitably and correctly defined or other appropriate information about the statistical significance of the experiments?
    \item[] Answer: \answerYes{}{} 
    \item[] Justification: We carry out significance testing as described in the methods (\cref{method}) section of the paper, for both alignment and downstream tasks. For our downstream tasks (\cref{fig:downstream_tasks_bar}), we also include error bars on the brain-tuned models. These are calculated using scipy.stats.sem on the metric scores across our 6 brain-tuned models. We do not include error bars in our brain-alignment plots (\cref{fig:brain_alignment_bar}) - as all values are paired within participants, conventional error bars for independent samples are not applicable. We perform a paired (wilcoxon) statistical test and indicate significance in the figure instead.

\item {\bf Experiments compute resources}
    \item[] Question: For each experiment, does the paper provide sufficient information on the computer resources (type of compute workers, memory, time of execution) needed to reproduce the experiments?
    \item[] Answer: \answerYes{} 
    \item[] Justification: At the end of \cref{subsec:braintuning}, we provide these details (Each brain-tuning uses 1 H100 GPU and 16 AMD EPYC 9654 CPUs on 244 GB of RAM, and takes approximately $70$ hours on an H100 GPU. Each evaluation uses the same compute resources, and takes approximately 90 minutes.) 
    
\item {\bf Code of ethics}
    \item[] Question: Does the research conducted in the paper conform, in every respect, with the NeurIPS Code of Ethics \url{https://neurips.cc/public/EthicsGuidelines}?
    \item[] Answer: \answerYes{} 
    \item[] Justification: We do not collect any data involving human participants, nor publish any novel dataset, nor expose any personal information. The CNeuromod dataset involving human subjects was collected with the approval the review board “CIUSS du centre-sud-de-l’île-de-Montréal” (under number CER VN 18-19-22). 

\item {\bf Broader impacts}
    \item[] Question: Does the paper discuss both potential positive societal impacts and negative societal impacts of the work performed?
    \item[] Answer: \answerYes{}
    \item[] Justification: In \cref{subsec:broader_impacts}, we discuss potential broader impacts of our work.
    
\item {\bf Safeguards}
    \item[] Question: Does the paper describe safeguards that have been put in place for responsible release of data or models that have a high risk for misuse (e.g., pretrained language models, image generators, or scraped datasets)?
    \item[] Answer: \answerNA{}{} 
    \item[] Justification: The paper poses no such risks (no novel data or models being released that have a possibility of misuse)

\item {\bf Licenses for existing assets}
    \item[] Question: Are the creators or original owners of assets (e.g., code, data, models), used in the paper, properly credited and are the license and terms of use explicitly mentioned and properly respected?
    \item[] Answer: \answerYes{} 
    \item[] Justification: In \cref{subsec:licenses} cite all existing assets used with their respective version.

\item {\bf New assets}
    \item[] Question: Are new assets introduced in the paper well documented and is the documentation provided alongside the assets?
    \item[] Answer: \answerYes{} 
    \item[] Justification: We release our code, alongside details about training and evaluation of our brain-tuned model see \cref{subsec:data_and_code_availability}.

\item {\bf Crowdsourcing and research with human subjects}
    \item[] Question: For crowdsourcing experiments and research with human subjects, does the paper include the full text of instructions given to participants and screenshots, if applicable, as well as details about compensation (if any)? 
    \item[] Answer: \answerYes{} 
    \item[] Justification: We include details of stimulus presentation as is relevant to the project (subjects watched Friends half-episodes), and don't include more detail as we feel it is not necessary to understand our method or results. Subject details can be found at \cref{appendix:participants}. Compensation information is not provided in the original cneuromod dataset. Further details can be found in the cneuromod documentation at \url{https://docs.cneuromod.ca/en/2020-alpha2/MRI.html#stimuli}

\item {\bf Institutional review board (IRB) approvals or equivalent for research with human subjects}
    \item[] Question: Does the paper describe potential risks incurred by study participants, whether such risks were disclosed to the subjects, and whether Institutional Review Board (IRB) approvals (or an equivalent approval/review based on the requirements of your country or institution) were obtained?
    \item[] Answer: \answerNA{} 
    \item[] Justification: The study did not itself collect any data from human subjects, but uses the cneuromod 2022-alpha release of the Friends dataset. All subjects provided informed consent to participate in the cneuromod data release, which was approved by the ethics review board of the “CIUSS du centre-sud-de-l’île-de-Montréal” (under number CER VN 18-19-22). For more details, see the cneuromod documentation: \url{https://docs.cneuromod.ca/en/latest/DATASETS.html#friends}

\item {\bf Declaration of LLM usage}
    \item[] Question: Does the paper describe the usage of LLMs if it is an important, original, or non-standard component of the core methods in this research? Note that if the LLM is used only for writing, editing, or formatting purposes and does not impact the core methodology, scientific rigorousness, or originality of the research, declaration is not required.
    \item[] Answer: \answerNA{} 
    \item[] Justification: Our research does not involve any LLMs as any important, original or non-standard components.
\end{enumerate}

\end{document}